\documentclass[letterpaper, 10 pt, conference]{ieeeconf}  %

\IEEEoverridecommandlockouts                              
\usepackage{graphics} 
\usepackage{epsfig} 
\usepackage{amsmath} 
\usepackage{amssymb}  
\usepackage{multirow}
\usepackage{float}
\usepackage{algpseudocode}
\usepackage{wrapfig}
\usepackage{subcaption}
\usepackage{pmboxdraw}
\usepackage{listings}
\usepackage{cite}
\usepackage{algorithm}

\usepackage{xcolor}
\usepackage{framed}

\makeatletter
\let\NAT@parse\undefined
\makeatother
\usepackage{hyperref}

\title{\LARGE \bf
ESNI: Domestic Robots Design for Elderly and Disabled People
}

\author{ 
   Junchi Chu \\
  \texttt{junchi\_chu@brown.edu} \\
  \texttt{Brown University} \\
  \and
  Xueyun Tang \\
  \texttt{xtang01@risd.edu} \\
    \texttt{Rhode Island School of Design} \\
}

\begin{document}

\maketitle
\thispagestyle{empty}
\pagestyle{empty}

\begin{figure*}[]
    \centering
    
    \includegraphics[width=\textwidth]{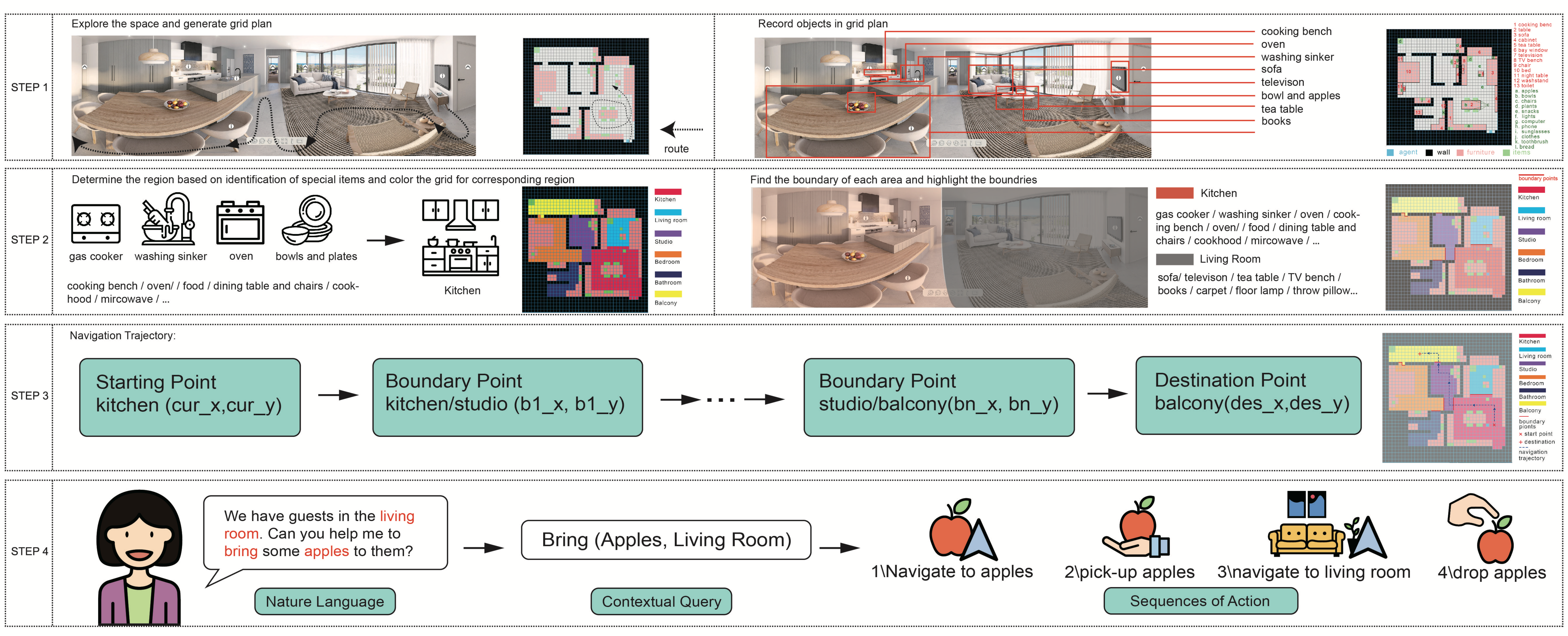}
    \\[\smallskipamount]
    
    \caption{Overview of the ESNI Design}
\end{figure*}
\begin{abstract}

Our paper focuses on the research of the possibility for speech recognition intelligent agents to assist the elderly and disabled people's lives, to improve their life quality by utilizing cutting-edge technologies. After researching the attitude of elderly and disabled people toward the household agent, we propose a design framework: ESNI(Exploration, Segmentation, Navigation, Instruction) that apply to mobile agent, achieve some functionalities such as processing human commands, picking up a specified object, and moving an object to another location. The agent starts the exploration in an unseen environment, stores each item's information in the grid cells to his memory and analyzes the corresponding features for each section. We divided our indoor environment into 6 sections: Kitchen, Living room, Bedroom, Studio, Bathroom, Balcony. The agent uses algorithms to assign sections for each grid cell then generates a navigation trajectory base on the section segmentation. When the user gives a command to the agent, feature words will be extracted and processed into a sequence of sub-tasks.    
\end{abstract}

\section{INTRODUCTION}


With the increasing aging population, the proportion of the global population over 60 years old is expected to reach 22 \% in 2050. Therefore, the changing of demographic has resulted in a lack of working populations, with a corresponding shortage in the human resource reserve of caregivers for the elderly and the disabled. Study shows that older people who live alone have the potential of health risks, such as joint disease puts them at higher risk of falls.\cite{kharicha2007health} Structural and functional impairments also affect the independent living of the disabled. The need of designing a household agent to improve the living standards of the elderly and the disabled have a significant project value. We assume that computer vision technology enables our agents to have the capacity of recognizing objects and the surrounding environment. The robots will be equipped with stereo cameras and various sensors to receive information and achieve autonomous navigation.

Based on research on the needs of the elderly and disabled group, our design focuses on household agents in the home environment. The agent follows our design procedure to generate a navigation system. Moreover, the agent receives voice commands from our target customers, maps natural languages to the contextual query language, and a sequence of actions.
The current state-of-the-art methods, use Simultaneous localization and mapping (SLAM)\cite{mur2015orb} as navigation algorithms, however, SLAM relies heavily on the landmarks and the poses of the robot during data acquisition. Occupancy grid mapping techniques indeed can partition the environment into cells for a better analysis, but the combined OGM and SLAM do not indicate the information of section segmentation and boundary points. The traditional path planning algorithm does not generate a navigation trajectory base on the human command directly\cite{976276}.

Thus, we address the above issues by introducing a navigation framework for agents in an unseen environment, to observe the grid cells, predict the section segmentation, discover all boundary points between each section, generate the section path at a high level and specify navigation trajectory details. As a result, the agent can process natural language to contextual query, then generate a sequence of the basic task to perform the corresponding assignments by using the navigation tools he learns.

\section{BACKGROUND}
\subsection{People's Attitude towards domestic robots}
Smart home products such as sweeping robots gradually become part of people's daily lives, which means the acceptance of home service robots' popularization. Voice interaction has become a trend for the current agent design principle.\cite{ray2008people}. In the questionnaire survey of disabled people and the elderly on home service robots, the respondents' needs for "use and maintenance of household appliances" and "delivery of goods" are relatively higher rather than other tasks\cite{harmo2005needs}. The survey mentioned a few most popular tasks that people expect the intelligent agent to achieve, such as "remind me to take medicine" and "grab me something I need". However, people still feel conservative about whether the agent can achieve tasks regards to some social activities and privacy, such as " playing cards like a real human" or "help me taking a bath".\cite{huang2021attitudes}.

\subsection{Contextual Query Language}

The contextual query language is a formal language that retrieves information in a well-defined structure\cite{hsiung2021generalizing}. The query, in general, must be human-readable and writable while maintaining the featured information of complex original languages. The meaning of using contextual query in this environment is to map from a border generalization of human instructions to a selection of a well-defined task. More specifically, we extract the verb and nouns words as keywords parameters in our model. 
\subsection{Occupancy grid mapping \& SLAM}

The occupancy grid mapping refers to a representation that the environment was partitioned into different cells. Each cell has a Bayesian probability to indicate whether it is 1(occupied) or 0 (empty). The representation is simply the situation of mobile robots and have become the dominant paradigm for embodied agent environment. \cite{4433772}

SLAM refers to Simultaneous Localization and Mapping, is an algorithm in which the agent starts the exploration from a random position and without any prior knowledge of the environment, and then calculates the relative position to the map and constructs the mapping of different areas. Kalman Filter Method is the most commonly used method in SLAM algorithms, which gives posterior pose estimation by using Bayesian filter. \cite{welch1995introduction}. 
\subsection{A* Algorithm}
A* Algorithm is a graph traversal and path finding computer algorithm that are widely used in games development. The advantage of A* algorithm is that, it can search a shortest path from source to destination with hindrances. A* algorithm selects the path with a minimum cost function f(n)=g(n)+h(n), which g(n) is the cost from source node to the current node, and h(n) is a heuristic function that estimate from the current point to the destination point. The selection of heuristic function depends on the problem, but in general, Manhattan distance or Euclidean distance are two most popular heuristic function to use. 
According to the above classification principles, the robot will predict the spatial type of each grid cell by calculating the distance of the observed objects in the space. For the space with key objects, if the agent observes a toilet accessible to a particular cell without colliding to a wall, then the robot can decide this grid cell belongs to the bathroom. For the space type without key objects, when the robot doesn't detect a key object at a certain grid, it will need to observe all objects in the space, and use the weighted sum of all items to calculate the probability of belonging sections\cite{duchovn2014path}. 
\subsection{POMDP}
Robots can be described as embodied agents either executing or learning a policy in an environment. This decision process can be formally described as a Markov Decision Process (MDP). An MDP is described by a tuple, $\langle\mathcal{S},\mathcal{A},T,R,\gamma\rangle$, where \textit{state space} $\mathcal{S}$ is the set of all possible states, \textit{action space} $\mathcal{A}$ is the set of all possible actions, \textit{transition function} $T=T(s,a,s')=P(s'|s, a)$ is a function that describes stochastic transitions from a state $s$ to a state $s'$ if executing action $a$, \textit{reward function} 

The agent may not be able to observe the exact state it is in, and instead can only make an \textit{observation} $o$ associated with the true state $s$. That is, the state $s$ is \textit{partially observable} by the agent as observation $o$. When states are partially observable, the MDP is termed a partially observable MDP (POMDP). POMDPs extend the MDP tuple with the addition of two elements $\mathcal{O}$ and $Z$: the \textit{observations space} $\mathcal{O}$ is the set of all possible observations that can be observed, and the probabilistic \textit{observation model} $Z=Z(s,o,a)$ describes the probability of obtaining observation $o$ \cite{zheng2021multi}.
\section{RELATED WORK}
The design of a household agent is a well-studied problem, particularly in an unseen environment. Charles suggests that the principles of designing an intelligent agent must be that the agent can make independent decisions rather than passive, and must be adaptive to the environment\cite{macal2005tutorial}. To consider the economic benefits of production, the agent has well-defined working domains, objectives, and pre-defined common rules. However, we identify that the main challenge is recognized as how to implement the path planning and navigation of the robot. Torvald proposed an online path-finding algorithm that solves the navigation problem in a maze environment\cite{ersson2001path}.  

Prior work uses deep reinforcement learning to explore the unknown environment for the embodied agent and defines an optimal policy in the standard Markov Decision Process framework to maximize the reward. For example, Bob and Marcin introduced a general approach to solving the exploration by using sparse rewards \cite{nair2018overcoming}. Iou-Jen contributed a cooperative multi-agent exploration (CMAE) that has shared goals for efficient exploration and learns coordinated exploration policies by using sparse rewards\cite{liu2021cooperative}. However, the main difficulty of turning a problem into MDP or solving an MDP problem, is that the state variables make the system complex as the real-world environment is constantly changing. thus,  it violates the assumption most MDP problems are discussed in a non-dynamic environment.

From a design perspective, a simplified environment can hardly compare to the real world, and thus must make strong assumptions that the design avoids the consistently changing physical issue. For example, when the wind knocks the book off the shelf. From the work done by John, the interface between the intelligent agent and environment need to be simple\cite{anderson1995generic}, because the generic simulator should control many aspects of the functionality of the environments. 

Our contributions differ from prior work in that we design a framework for an embodied agent that is applicable for any house plan in the real world. The agent can learn the surrounding without supervision, and autonomously build its knowledge based on the algorithm we implement. The agent can understand human command and perform simple actions since it can move freely in the grid world by planning the navigation trajectory.

\section{APPROACH}

\subsection{Jargon}
\subsubsection{Symbols}
Before elaborating on our methodologies to address the tasks in this project, we first 
give some symbols and terminology definitions as follows.
We denote the collections of all moveable objects in the grid world as M = \{$m_1$,$m_2$,$m_3$,...,$m_n$\}. The target grid's coordinates are denoted as ($grid_x$,$grid_y$) and the target object mj's coordinates are denoted as $(mj_x,mj_y)$. All unmovable objects,(e.g., walls and furniture ), are defined as U=\{$u_1$,$u_2$,...,$u_n$\}. The agent's current location is denoted as $(cur_x,cur_y)$. 
\subsubsection{Definitions}

\paragraph{Sections}

The collection of sections is denoted as: 
$\mathcal{S}$=\{$s_1$,$s_2$,...,$s_x$\}
For each grid cell $c$, a section $s_j,j<=x$ is an accumulated grid cells cluster, $c\in\mathcal{S}$, that can represent the functionality of a house, for example, kitchen, living room, bathroom, etc. 
\paragraph{Boundary Points}
The collections of points that lays between two different sections. For any point is identified as a boundary point bn, there exists a neighbour point belongs to a section that is different from bn's section. Mathematically, Suppose a boundary point is $(bn_x,bn_y)$, we have $\forall  (bn_x,bn_y)\in s\_p$ , $\exists  (bn_x+a,bn_y+b)\in s\_q $, such that  $a,b \in$ \{-1,0,1\} , $a+b\in$ \{-1,1\} and $p \neq q$.
                     
\paragraph{Section Path Finding}

Section path is a path from a section to another section. The purpose of section path finding is to find a path in high-level to help the agent's navigation trajectory. For example, to navigate from a point in kitchen to bathroom, the section path is:  Kitchen $\rightarrow$ Studio $\rightarrow$ Bedroom $\rightarrow$ Bathroom . The section path can be generated by the BFS algorithm discussed in the section segmentation paragraph. 

\paragraph{Walking Values in Exploration}

Walking Value is a term to describe the frequency of the agent visit on a particular grid. We create a dictionary that stores the information of each grid to associate a walking value. We initiate all walking value to be 0, and increase the value by one when the agent visit that position. 

\paragraph{Histogram Voting System \& Bin Values}

Histogram Voting System is an algorithm for section segmentation. The agent creates a histogram to estimate the likelihood or to vote for the assignment for a particular grid. The bars of a histogram are called bins, and we have 6 bins in total for the histogram since there are 6 sections in the grid world. Each bin has a numerical value as Bin Values, we assign a grid to a section if it has the highest bin value associated with the section.

\paragraph{Ground Truth Knowledge}

The knowledge we gave to the agent was that for each item he observes, what section is most likely to contain that item.

\subsection{Agent Model and Setup}
We converted the architectural floor plans into the digital 40 x 40 grid world by using spatial mapping. We defined a white area as a walkable space for the agent, and any space other than white is non-walkable.  
The initial position of the agent is located at the right corner of the building, with no prior knowledge of the environment. When initializing the environment, we instantiate objects in the zones by sampling the attribute distributions in the knowledge dictionary (explained), which captures an ontology of locations and objects in the building. Objects are classified as moveable objects (such as apples, bananas, computers, etc.) and unmovable objects (such as walls, beds, dining tables, etc). The goal is to explore the room randomly with efficiency in the stimulating environment, and we assume the agent can recognize objects with high accuracy by using computer vision techniques, that is, when the agent encounters an object, the attributes of the item need to be scanned in agent’s memory and only move in the walkable space (white grids).
\begin{figure}

\includegraphics[scale=0.078]{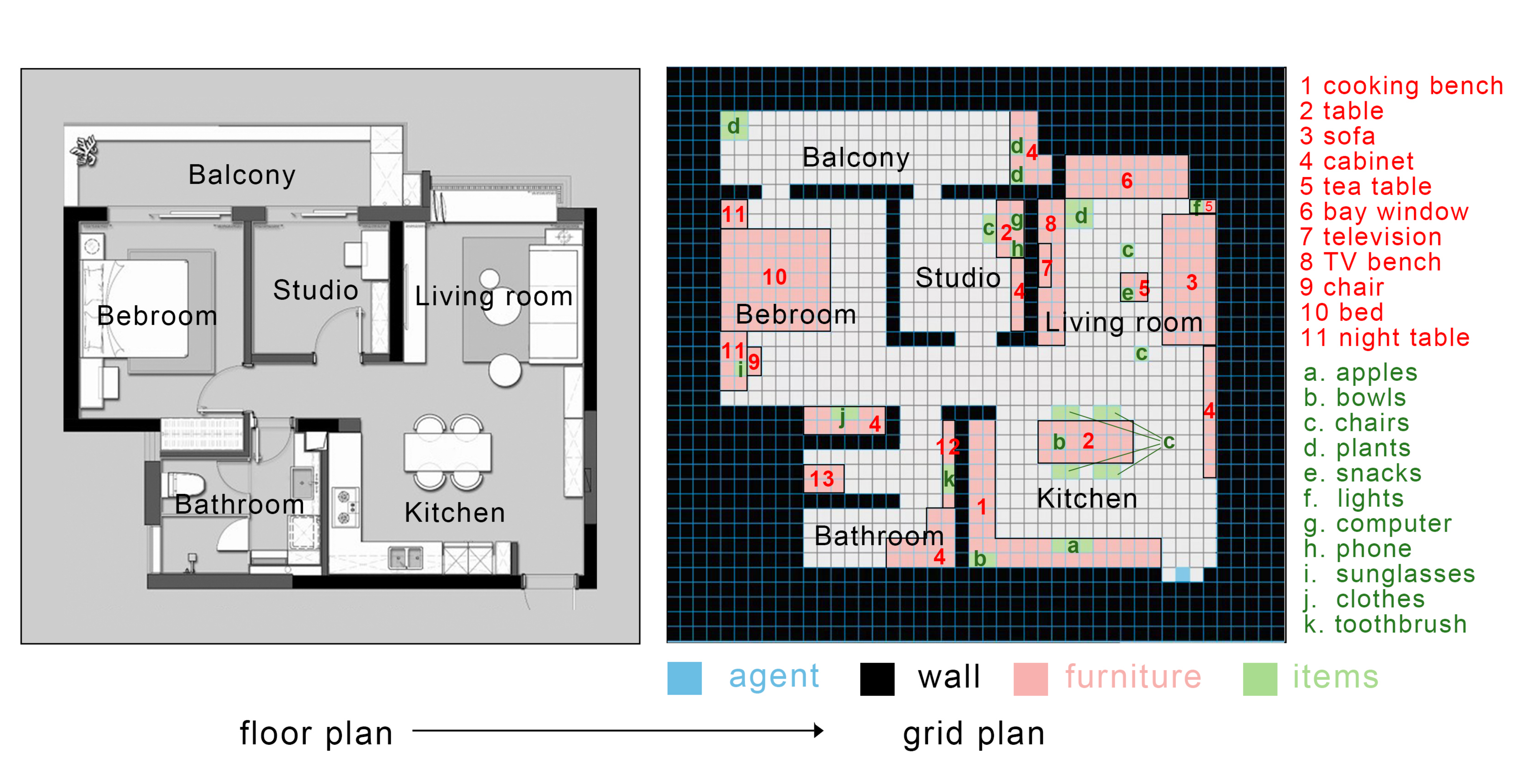}

    \caption{A Partition From House Plan to Grid Cells}
\end{figure}

\subsection{Exploration}

In this project, we assume that the agent can perceive objects are close to them. In the 2-d grid world, we use Euclidean distance to calculate the diagonal distance between objects and the agent. Even though the computer program has the exact information of all objects, the agent does not. When the agent is nearby the object, we add the object property, including object name, ID and attributes to the agent's memory. The goal in this step is for the agent to cover as many unvisited spaces as possible, identify all the boundary points, and build the unweighted connectivity graph to develop a path finding algorithm, efficiently.

\begin{algorithm}
\caption{Agent Exploration Algorithm}\label{alg:cap}
\begin{algorithmic}[1]
\State Assume agent's current positions are $(cur_x,cur_y)$
\State Initiate a dictionary: walking-values[$(cur_x,cur_y)$] = 0
\While{min(walking-values.getValues()) $!=$ 0}
\State MOVES $\gets$ ["UP","DOWN","LEFT","RIGHT"]
\State Remove move if collides with obstacles 

\For{each move in MOVES}
   \State move = argmin(walking-values[move])
\EndFor
\State walking-values[$(cur_x,cur_y)$]+=1
\EndWhile
\end{algorithmic}
\end{algorithm}

The algorithm can be described as follows: we create a dictionary WALKING\_VALUES that map from all positional and walkable grids to a numerical value. We mark a grid that has been visited by increasing the value by 1, if the agent has stepped over or passed that grid. In this simulated environment, the agent has 4 legal options for the next move, UP, LEFT, DOWN, RIGHT. But the agent should not collide with any obstacles, so we remove the potential choice of the next move if the agent will collide with an object or obstacles. Based on 4 possible move options, the agent chooses the move which has the minimum walking value in the dictionary. Intuitively, if a move directs the agent to a grid that has a higher walking values than others, that means that particular grid is more frequently visited than others, then the agent will choose the next move of the grid with a lowest value among the 4 directions,

\subsection{Section Segmentation}

\subsubsection{Section Assignment}

The agent needs to have an understanding of the current environment, meaning a human will not manually indicate each grid. We design a section prediction algorithm that helps the agent can guess the area correspondingly to distinguish the different functionalities of areas (such as balcony, studio, kitchen, etc). After agents explored all spaces in the grid world, all item information must be recorded and the next procedure for the agent is to predict areas for each grid. The agent uses an area prediction algorithm to determine which area a grid is belonged to. 

To help the robot perform the section segmentation, which is assigning a section to each grid cell, we divided six basic types according to the functions of the home space: living room, bedroom, bathroom, studio, kitchen, and balcony. We cluster the bedroom, bathroom, and kitchen into one group with the identification of key objects, and the rest to be the other group without key objects. Key object refers to a signature object that determines the functionality of a section, for example, the bathroom has a toilet as a key object, a space with the bed must be the bedroom, and space with a gas cooker must in the kitchen. The toilet, bed, and gas cooker can be regarded as the key objects for deciding the space types. However, for the other three types: Living room, Studio, and Balcony, lack of key objects needs us to use other objects to measure the space types. Figure 3 is a table of 2 groups of space types and with association of key objects.  

According to the above classification principles, the robot will predict the spatial type of each grid cell by calculating the distance of the observed objects in the space. For the space with key objects, if the agent observes a toilet accessible to a particular cell without colliding to a wall, then the robot can decide this grid cell belongs to the bathroom. For the space type without key objects, when the robot doesn't detect a key object at a certain grid, it will need to observe all objects in the space, and use the weighted sum of all items to calculate the probability of belonging sections.  


\begin{figure}[h]
\includegraphics[scale=0.2]{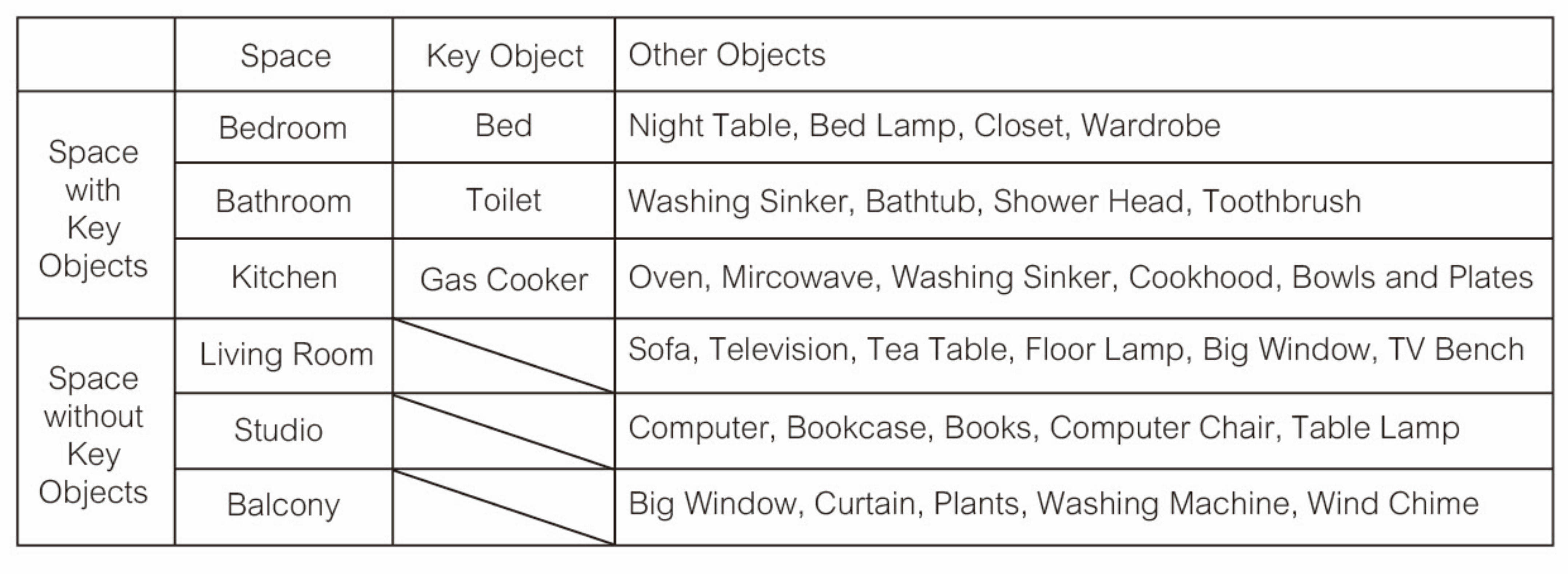}
    \caption{Table of identifying key objects and other objects}
\end{figure}




There are some edge cases that need to be discussed. For example In this case, for these grids in the bathroom area, toothbrush and toilet are the only 2 items that can contribute weights to the histogram bins. But there are more than 8 objects in total located in the kitchen that can affect the result of the histogram so that the agent might predict some grids in the bathroom as kitchen, which is obviously wrong. To solve this issue, we significantly scale down the contribution of bins value for each object by a discount factor $N$, if the diagonal between the object and grid encounters wall objects. Thus, grids in the bathroom area but near the left side of the kitchen near the wall are classified properly.

The overview of the area prediction algorithms is described as follows. Previously, the agent explored all sections and stored the location information for each item. For each grid, 6 sections correspond to 6 possible assignments, thus we create 6 bins in the histogram. The agent calculates the Euclidean distance between each item in his knowledge and for each item associated with the zone. We sum up the reciprocal of Euclidean distance by times with a factor N, to the associated bin to account for how possibly this grid belongs to the zone.Mathematically, the histogram formula can be represented as the follow:
\[ Bin\_Value =\sum_{j=1}^{j=n} \frac{1}{N((grid\_x-mj\_x)^2+(grid\_y-mj\_y)^2)} \]

\begin{algorithm}
\caption{Agent Section Segmentation Algorithm}\label{alg:cap}
\begin{algorithmic}[1]
\While{Exist a grid without Section Segmentation}
\State Initiate a histogram: Hist\_Area that has 6 bins:Balcony, Living Room, Bedroom, Bathroom, Kitchen, Studio 
\For{each object o in Grid World}
\State Calculate Bin\_Values
\For{each bin in Hist\_Area}
\If{bin in Ground Truth\_Knowledge[o]}
   \State bin += Bin\_Values 
   \EndIf
\EndFor
\EndFor
\State Area = argmax(Hist\_Area)
\EndWhile
\end{algorithmic}
\end{algorithm}
\begin{figure}[H]
\includegraphics[scale=0.1]{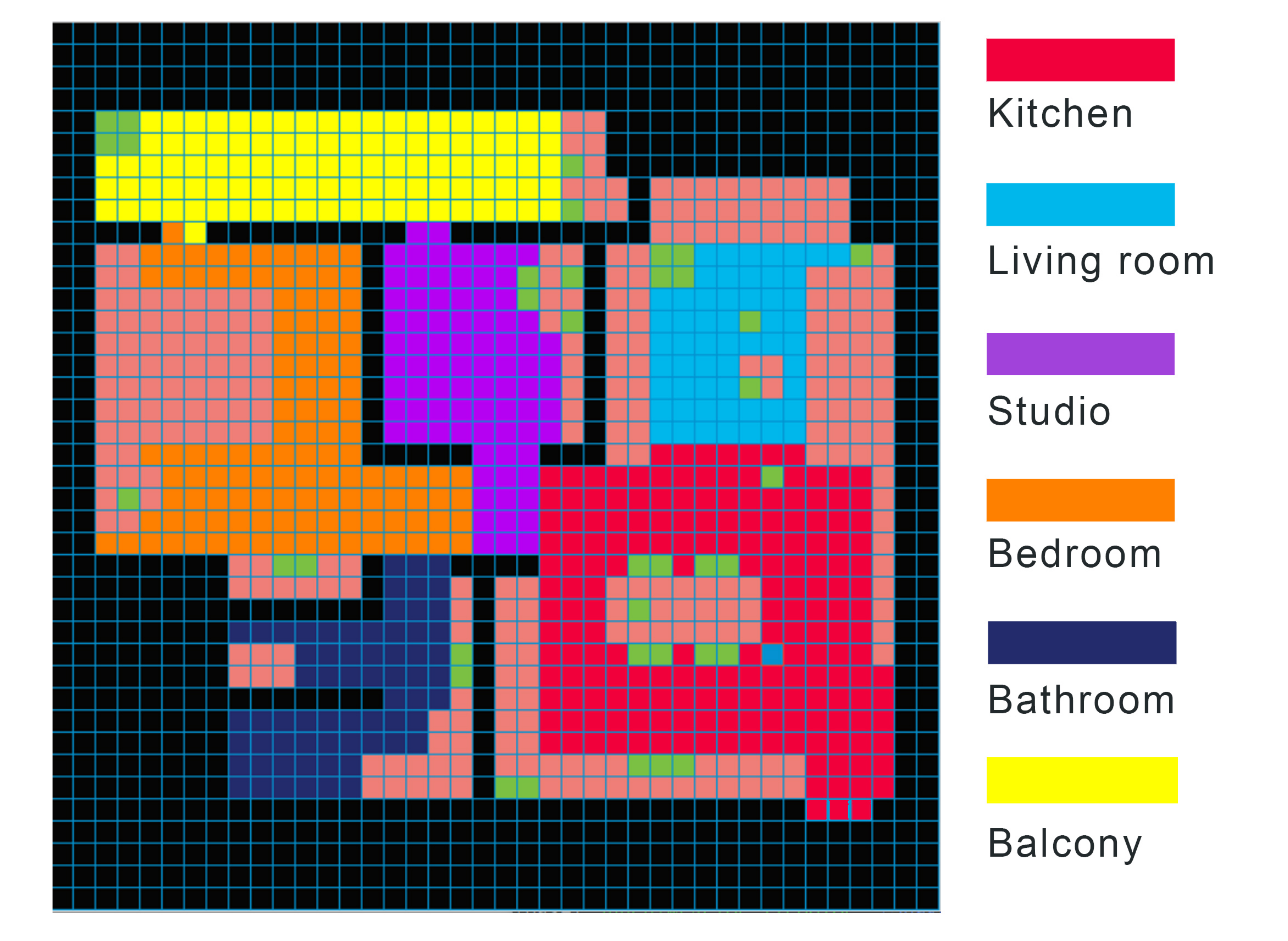}
    \caption{The result of Section Segmentation: Paint Sections to Different Colors}
\end{figure}

\subsubsection{Boundary Points}
Our goal is to design navigation algorithms that are suitable for all kinds of homes, and we make the assumption that the agent does not hold any knowledge of taking shortcuts from a start point x to a destination point y. 
The agent uses the section prediction algorithm to predict belonging sections for each grid, to set up the navigation system, it needs to find all the boundary points. We define a boundary point, in this simulated environment, a point that belongs to section A but any adjacent point that is predicted to a section rather than section A, then we consider this point as a boundary point. The collection of boundary points is useful when we build the navigation system for the agents

\subsubsection{Section Path Finding}
In the grid world, the exploration helps to build an unweighted graph into the agent’s memory. The graph instructs the agent to detect a path from its current location to the destination point in the level of sections. In our floor plans, the balcony is adjacent to the bedroom and studio, but not near to the living room. Hence, to navigate from the balcony to the living room, the agent must bypass the studio and kitchen, then reach the destination. 
Section path finding can be achieved easily by using a  deterministic algorithm (Breadth-First Search) \cite{bundy1984breadth} to determine the shortest path from the current location section to the destination section. The unweighted graph in general is not complex, so the BFS traversal can provide us satisfying results in terms of the connectivity of different sections.  

\begin{figure}[h]
\includegraphics[scale=0.45]{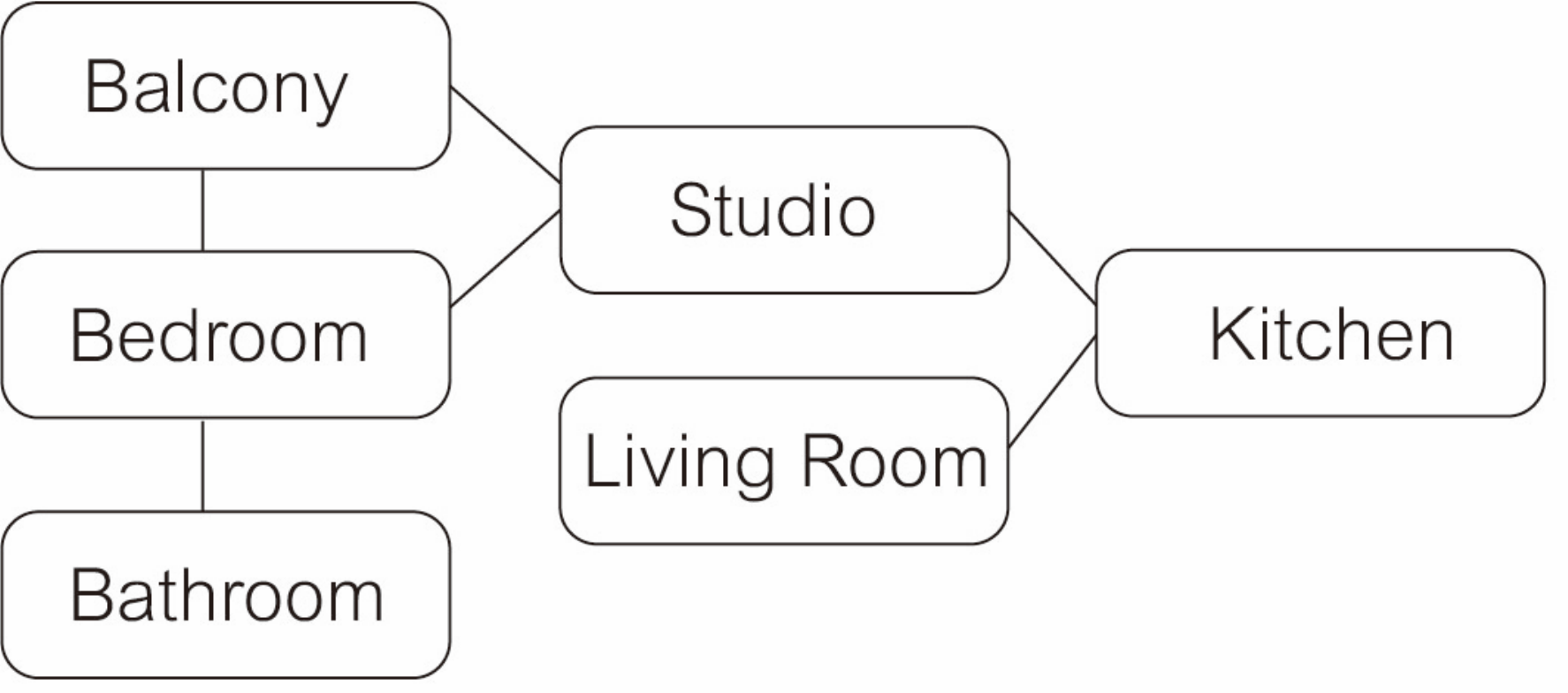}
    \caption{Section Connectivity Unweighted Graph}
\end{figure}

\subsection{Navigation Trajectory}

Given a destination point ($des_x,des_y$) and the agent’s current coordinates ($cur_x, cur_y$), the agent uses the section segmentation function to locate its current section and destination section. From the unweighted connectivity graph for sections, the BFS algorithm generate the path. Between those areas are called intermediate areas, and the navigation strategy for the agent is to move to the boundary points for the current area and the next intermediate area, according to the high-level path from BFS. Once the agent reaches the boundary point, it needs to navigate to the next boundary point between intermediate areas, until it reached the destination section. 

For example, consider the following scenario, the agent is currently located in the bedroom. We send out a command that requests it to carry an apple to the people in the balcony. The natural language could read as: "There are guests in the balcony, can you carry some fruits to serve our guests?". In the section levels, the navigation pipeline for the agent is: from bedroom, go to kitchen (carry an apple), go to studio, go to balcony. At the coordinates levels, the trajectory can be expressed as follows: agent's current location($cur_x$,$cur_y$) $\rightarrow$ boundary point between bedroom and kitchen ($b1_x$,$b1_y$) $\rightarrow$  apple's location ($apple_x$,$apple_y$) $\rightarrow$  boundary point between kitchen and studio ($b2_x$,$b2_y$) $\rightarrow$  boundary point between studio and balcony ($b3_x$,$b3_y$). The navigation between each step can be simply achieved by using the Manhattan path algorithm.  

\begin{figure}
\includegraphics[scale=0.75]{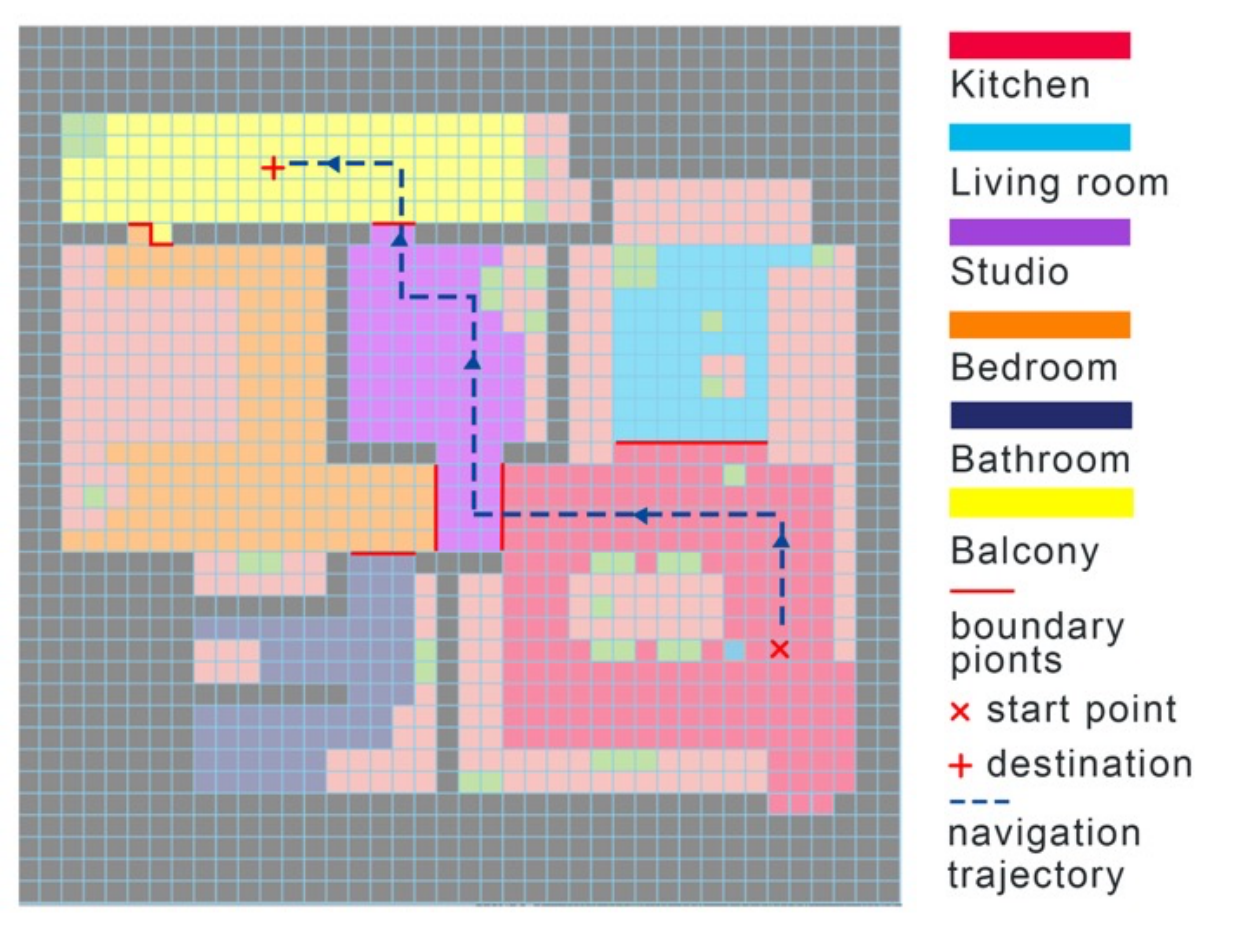}
    \caption{Blue Arrows Indicates a Navigation Trajectory From a Starting Point at Kitchen to the Destination Point at Balcony}
\end{figure}
\subsection{From NLP to Contextual Query}
The contextual query is a formal language for representing queries in a structured manner. Each class of contextual query corresponds with a task parameterized by a goal. 

The agent parses the contextual query and maps it to a valid sequence of parameterized sub-tasks. We use NLTK \cite{loper2002nltk}  to tokenize and tag the queries, then analyze the sentence composition to extract the goal parameters. We implemented custom transformations mapping from each contextual query class to goal-parameterized tasks. We use a natural language classifier to classify a natural language query  into the corresponding contextual query via the semantics mapping.

We define 4 types of contextual queries in this environment: "bring", "navigate", "find" and "swap". NLTK extracts the keywords in the natural language command and passes keywords as parameters to locate the specific contextual query. There are in total 4 basic tasks: navigate an object, pickup an object, drop an object, navigate to a section. Each contextual query is associated with a sequence of basic tasks.

\begin{figure}[h]
\begin{framed}\tiny
\text{Examples of mapping between NLP $\rightarrow$ CQ} 
\begin{enumerate}[]
\item\begin{itemize}
    \item \texttt{Task 1:I want an banana. I am at bedroom} Bring [Banana, Bedroom]
    \item \texttt{Task 2:Can you come to my bedroom to serve me?} Navigate [Bedroom]
    \item \texttt{Task 3:Hey, where is my computer? I can't find it.} Find [Computer]
     \item \texttt{Task 4:Hey, I want to take a shower. Can you swap my cloth and toothbrush? } Swap [Cloth, Toothbrush] 
    
\end{itemize}
\text{CQ $\rightarrow$ Sequences of Action} 
\item\begin{itemize}
    \item Bring[Banana,Bedroom] \textit{ find(Banana)$\rightarrow$navigate(Banana)$\rightarrow$pickup(Banana)\\$\rightarrow$navigate(Bedroom)$\rightarrow$drop(Banana)}
   
     \item Navigate[Bedroom] \textit{navigate(Bedroom)}
  
    \item Find [Computer] \textit{navigate(Computer)}
                
      \item Swap[Cloth,Toothbrush] \textit{ find(Cloth)$\rightarrow$navigate(Cloth)$\rightarrow$pickup(Cloth)\\$\rightarrow$find(Toothbrush)
      $\rightarrow$navigate(Toothbrush)$\rightarrow$drop(Cloth)$\rightarrow$pickup(Toothbrush)\\$\rightarrow$navigate(Cloth origin)$\rightarrow$drop(Toothbrush)}
\end{itemize}
\end{enumerate}\begin{flushleft}Figure \ref{fig:query_classes}: (a) Classifying natural language queries into parameterized contextual queries. (b) From contextual queries to a sequence of actions\end{flushleft}\end{framed}
    \phantomcaption\label{fig:query_classes}
\end{figure}

\section{Experiments}

Our experiment demonstrates the trade-off between steps taken by the agent and the sections he is able to explore in the grid world. The simulated grid world is relatively small but in the real world example, the space might be much larger. Theoretically, with limited steps the agent might not fully explore the grid world, thus will result in an incorrect section prediction. The maximum allowed step (MAS) is passed as an input for each iteration. The agent stops the exploration and performs the section segmentation once the maximum allowed steps are reached. For the exploration experiment, the agent starts at MAS of 50, and we increase a value of 50 steps to MAS, until 2000. We take the average results of 20 iterations for each epoch of MAS. For the section segmentation experiment, we use the same measurement metric as the exploration experiment, to get the results of section segmentation in different values of MAS.

\section{RESULTS}

\subsection{Results for Exploration}
When the maximum allowed steps are relatively low, the agent hovers around in those areas near the starting point, thus will prediction sections to only limited categories. By the mathematical nature of our exploration algorithm, if we increase the maximum allowed steps, the agent should ultimately step in other sections that he has not previously covered. A percentage of covered space with respect to maximum allowed steps can demonstrate the result of exploration. Fig.8 shows the percentage that the agent has covered or explored, given the maximum allowed steps. To achieve approximately 80\% explored space in the grid cells, the agent has a maximum allowed steps of 750. The curve converges to 100\%, which means that the agent has fully explored all the grids in this environment when the maximum allowed steps are around 1600 steps. 
\begin{figure}[h]
\includegraphics[scale=0.6]{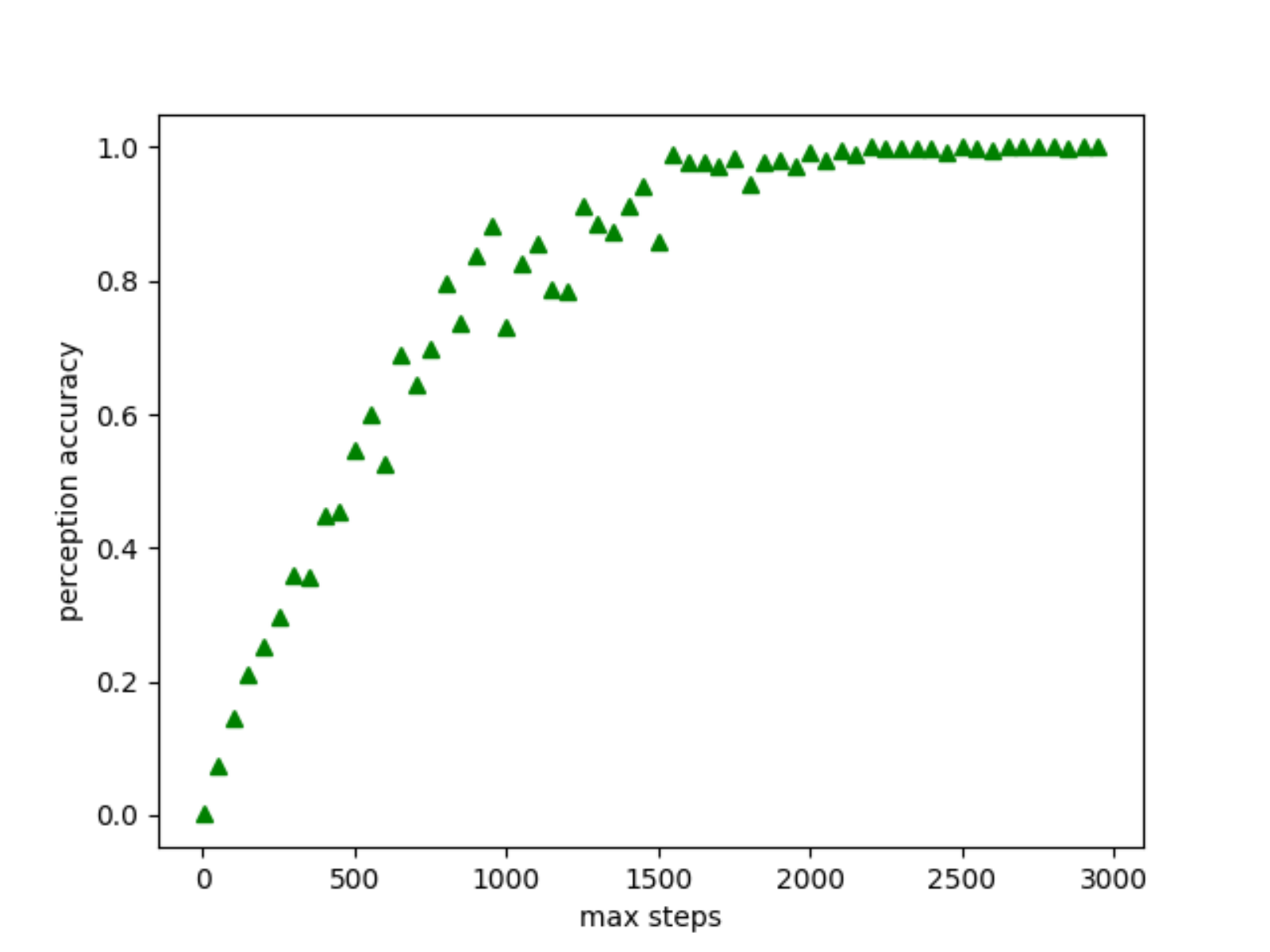}
    \caption{Demonstrate the correlation between MAS and space explored by the agent}
\end{figure}

\subsection{Results for Section Segmentation}

The number of sections for the agent can predict is determined by how many steps the agent can explore. Theoretically, all sections are segmented well if the agent can fully explore the grid cells and observe all items in the environment. Based on the average results of 20 experiments, Fig. 10 shows the number of sections the agent can recognize, given an assigned maximum allowed steps. 

\begin{figure}[h]
\includegraphics[scale=0.25]{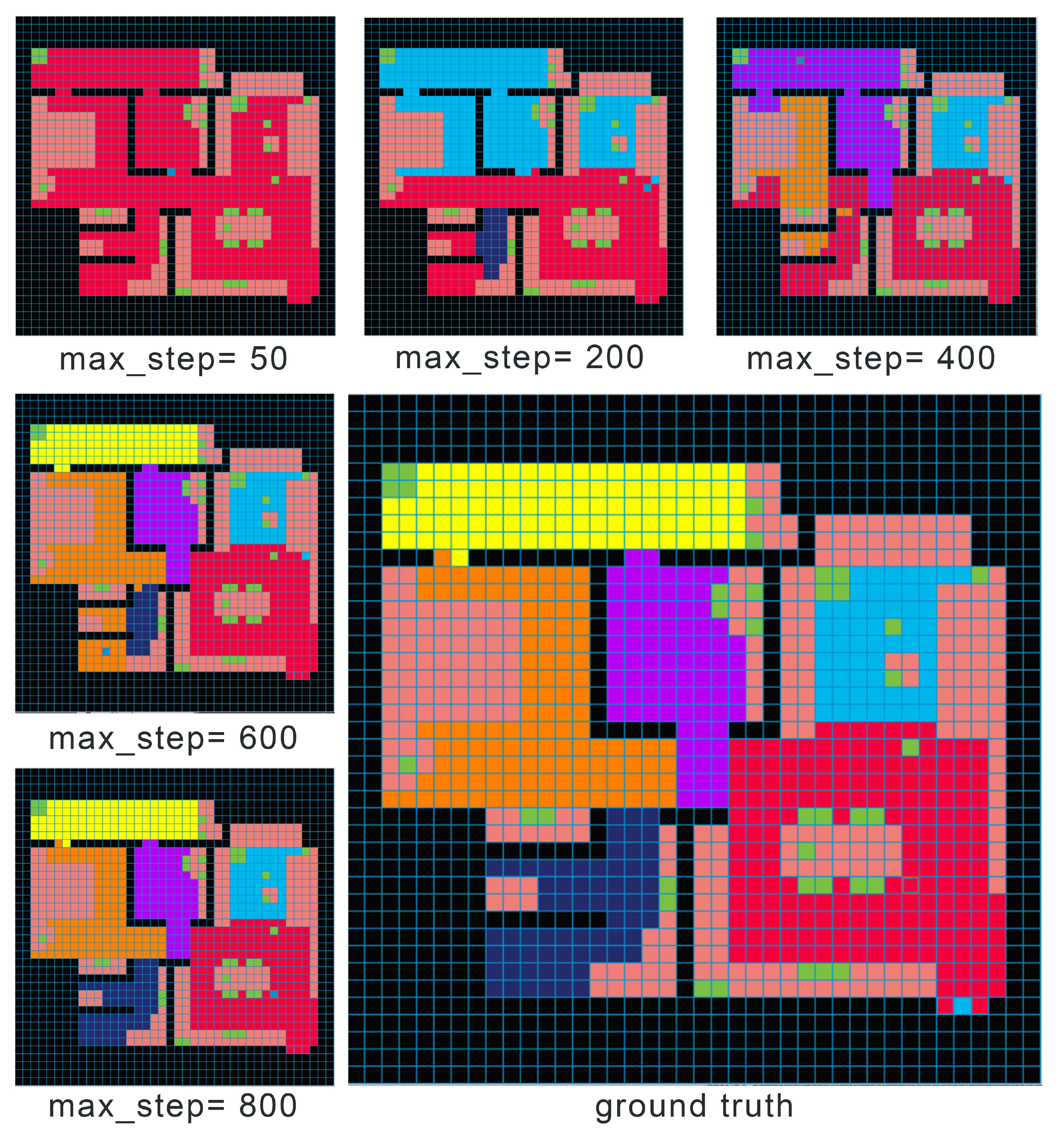}
    \caption{The result of section segmentation for different MAS in the computer program visualization.}
\end{figure}

\begin{figure}[h]
\includegraphics[scale=0.3]{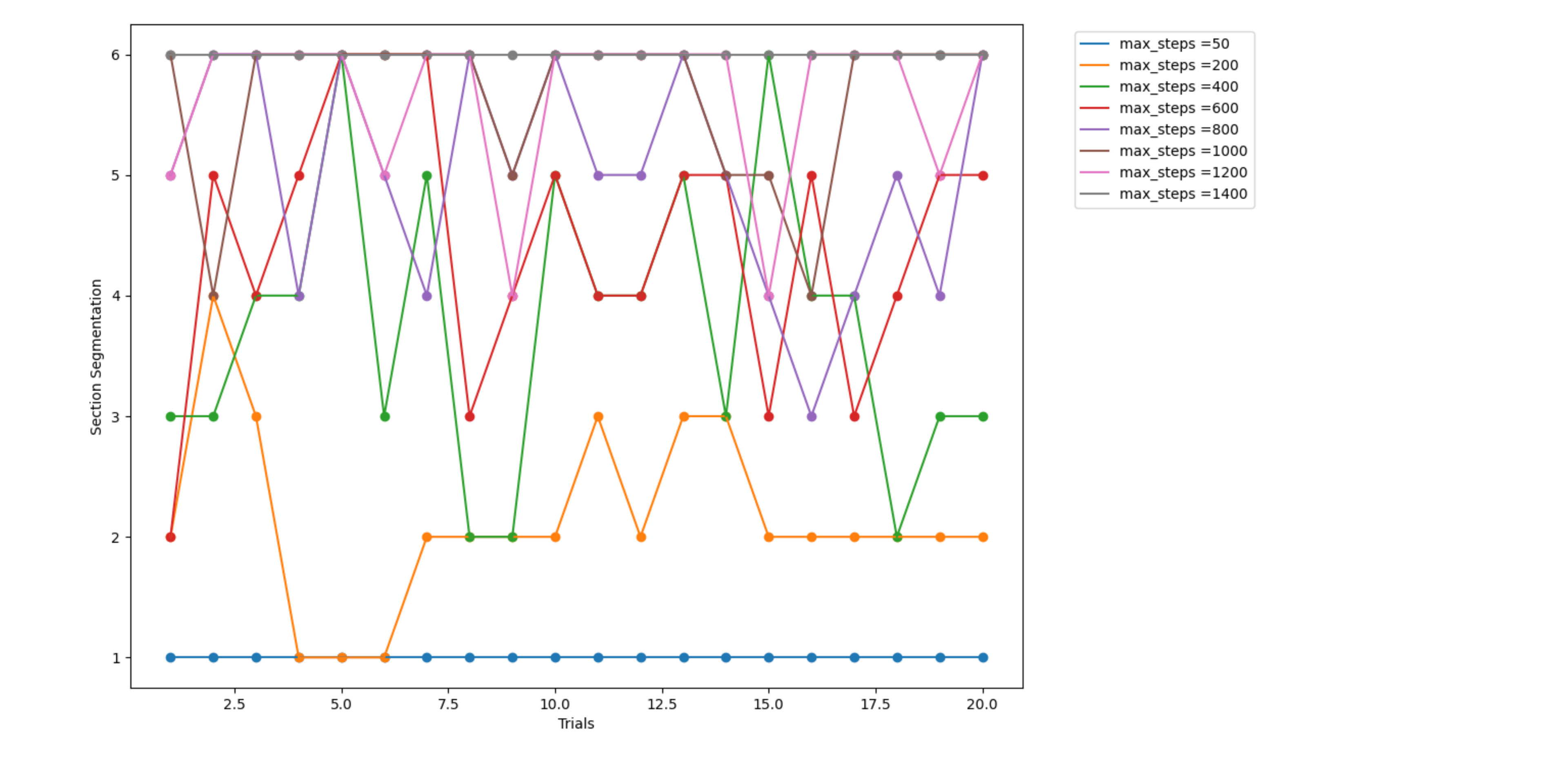}
    \caption{20 trials experiments' section recognized on different MAS }
\end{figure}

\section{DISCUSSION \& FUTURE WORK}

We present a household robot system for people with disabilities. The method is general and can be applied to different home plans. The agent does not necessarily need prior knowledge to develop its own navigation system but it does take steps to explore the unknown environment at first.

One of the limitations of this work is that our project relies on the assumption that the agent can move freely in the grid world, not considering the physics of the agent, but in reality, the agent can collide with obstacles that are never seen before in the memory. Another limitation is that the navigation algorithm suffered from the model of grid cells that do not have well-defined sections, for instance, a maze world environment.   

Future work can focus on the implementation in a 3-D environment, to have a system of complexity in vertical levels for the agent to interact with objects in the grid world. AI2-THOR framework \cite{kolve2017ai2} provides free online environments for various house plans that can be met to ascertain the value of this project, There will be more challenges when we transfer the experiment domain from 2-D to 3-D spaces, but the research on the design of intelligent agents has a profound impact commercially. 

\section{CONCLUSION}

We introduce a framework of intelligent agent design and provided several algorithms during the pipeline for the agent to perform tasks. The framework has novel points of discovering hidden information during the section segmentation and generating a navigation trajectory. The autonomous agent process human languages and utilize the trajectory to perform several actions. Our experiments show that our approach achieves satisfactory results compared to the state-of-art SLAM approach. Finally, we discussed the limitations of our approach and the extended research in the relative areas.

\small
\bibliographystyle{IEEEtran}
\bibliography{sample}

\end{document}